# Temporal-related convolutional-Restricted-Boltzmann-Machine capable of learning relational order via reinforcement learning procedure

Zizhuang Wang

*Abstract*— **Recent works on recurrent neural network and deep learning architecture have shown the power of deep learning in modeling time dependent input sequences. Specific learning structure such as Higher-order boltzmann machine, gradient-based learning manifold, and Recurrent "grammar cell" reveal their ability to learn feature transformation between related input maps, and perform well in time-related learning & prediction tasks in higher order cases. In this article, we extend the conventional convolutional-Restricted-Boltzmann-Machine to learn highly abstract features among abitrary number of time related input maps by constructing a layer of multiplicative units, which capture the relations among inputs. In some cases, we only care about how one map transforms into another, so the multiplicative unit takes features from only this two maps. In other cases, however, more than two maps are strongly related, so it is reasonable to make multiplicative unit learn relations among more input maps, in other words, to find the optimal relational-order(number of related input maps that the multiplicative unit extracts features from) of each unit. In order to enable our machine to learn relational order, we developed a reinforcement-learning method whose optimality is proven to train the network.**

## Ⅰ. Introduction

Unsupervised learning combined with deep architecture has unveiled part of the mystery of artificial intelligence. Such learning techniques have boarder applications in areas like visual recognition, natural language processing, audio detection, and cognitive analysis. With growing computational capabilities, deep learning framework like convolutional deep belief network **[1]** can bring more contribution to cognitive science. Recently, researchers begin to take the next step, trying to develop model that can handle time dependent learning tasks. Traditionally, recurrent neural network(RNN) has shown its efficiency in time-dependent recognition problems. For instance, RNN is widely used as an elegant framework to manipulate audio model **[2]**, which is based on time-related inputs. Recent researches also combined RNN with the power of convolutional restricted boltzmann machine(CRBM), such as gated autocoder and factorized CRBM **[3][4]**.

Later, more attention has been driven to a special form of time-based model, in which a restricted boltzmann machine is built to learn feature transformations that describe relations between time-related input maps. In the model, which is similar to conventional RBM **[5][6]**, a hidden stack of layers is constructed to describe conditional probabilistic distributions over inputs. The difference from traditional model is that the hidden layers take into account of several input maps at different time moment. That is, the hidden layers are able to extract features from a combination of several time-related maps. By doing so, we can extract features that are explained by hidden layers to represent correlation between inputs maps, in other words, to describe a matrix transformation from one map to another. Previous works **[7][8][9][10]** noted that multiplicative interaction is an effective way to correlate input maps. We use this method in our model to combine related inputs. With the learned hidden features, or transformation, our machine can predict later inputs based on conditional distributions that are learned and carried by the hidden layers of RBM. Also, with the power of high-order temporal dependencies that is describe by **[10]**, we can learn features that are even more abstract. In orther words, we can learn the features of transformations of input maps. This can be achieved with more hidden layers to be constructed and through learning efficiency brought by deep learning architecture.

There are limitations in higher order temporal model. Since conventional multiplicative interaction only takes into account of
two related maps, it lacks the ability to correlate
more input maps (input maps sequence for instance) and therefore can only learn features from two related inputs. Theoretically, we can learn temporal
dependence among inputs no matter how far they are through high-order training process. Thus, combining only two input maps through multiplicative interaction seems to be achievable and efficient. In practice, however, this will cause the number of layers and parameters that are needed to be learned to explode. Moreover, in some cases in which we may have a large amount of similar or strongly correlated input maps sequence, it is wise to combine them all together and to use only one stack of hidden layers to describe their correlation, saving a lot of space for parameters and time for learning. To do this, we define the term "relational order" as the number of maps that one stack of hidden layers learn features from, in other words, the number of maps that we correlate with each other using multiplicative interactions. Finally, we developed a reinforcement-learning **[11]** based method to learn the relational order at each time by minimizing reconstruction error. We then proved that it satisfies the sub-problem structure of dynamic programming **[12][13][14]**. Therefore, by finding the optimized reconstruction error at each group of related input maps, we get the globally optimal solution of the entire input sequence.

## II. RELATED WORK

Traditional temporal-dependent RBM is widely used to model language processing, audio recognition and other time-evolutional learning tasks . In **[15][16]**, deep belief network has revealed advantages over Gaussian mixture models for automatic speech recognition. Recent works **[17]** in image recognition show that a convolutional network (CNN) can dramatically enhance the power of artificial neural network. Basically, a CNN architecture includes hidden layers that are gained from convoluting high dimensional inputs with kernels, and a pooling layer to exclude noise in order to gain highly abstract features. By constructing a convolutional layer, CNN takes the advantage of local invariance and parameter sharing **[18]**. In image classification tasks, since the objects that need to be recognized from the image input may show up at different locations and angles, it is crucial to equip the model with the translational invariance property. By sharing parameters, on the other hand, we reduce parametric redundancy and therefore save a lot of space. Thus, instead of defining independent hidden units learning features in different parts of the input space, we make hidden units share the same combination of weights to extract features that may come up from different locations. These advantages make CNN extremely efficient. Therefore in this article, we developed a model based on convolutional deep neural network that takes temporal dependency into account. Such model combines the power of deep convolution and the capability of time-based model that can handle temporal related inputs. We showed its efficiency in tasks of learning matrix transformations of related inputs.

## III. MODEL

Recently, convolutional Restricted Boltzmann Machine has been used to extract features from high dimensional and highly abstract dataset **[1]**. In case of image processing, we extract features in the image by convoluting it with a 2d kernels and then construct a 2d layer of RBM **[1]**. After updating the kernels using CD-k algorithm, we are able to perform the task of image recognition efficiently by sending the outputs to a classical discriministic layer, or to generate an image map sampling from the Joint distribution P(X,Z), where the latent variable Z is described by hidden layers of C-RBM **[5][6]**.

Conventional recurrent network(RNN) has been proven to be a useful model that can handle time dependent learning tasks such as recognition and prediction for future outcomes based on previous inputs. In order to combine the power of deep learning and the ability to extract features from time sequence, researchers have developed structures that collaborate RBM with RNN. In **[3]**, for example, stacks of hidden units have been used to model time evolution features, and then a hidden pooling layer computes the sum of these hidden units by a sigmoid function, and finally update the weights by minimizing the distance function between the real visible inputs and the generative visible variables sampled from conditional distribution of P(v|h). Such model provides the basic architecture for audio recognition and other time-evolved learning tasks.

However, unlike the model of deep audio recognition framework, in which the input is a 1d time-related sequence and the machine constructs a distribution over hidden layer to capture time dependent features of sequenced audio signals, the 2d C-RBM lacks the ability to grab relations between time related sequence of images. Therefore In this article, we use a different framework, which combines the structure of high dimensional convolutional-neural-network with restricted boltzmann machine by constructing only one stack of hidden variables. The idea is, instead of building stacks of hidden layers, that the model constructs multiplicative units relating the present input with the previous ones. In this model, hidden variables are viewed as a layer that portrays correlation among observations at different time t. For prediction, the hidden layer generates future maps based on the learned probabilistic distribution. In many cases (Bi-Temporal-related model, Section 4), each multiplicative unit takes account of two input maps. One major drawback of this framework, therefore, is that the model only takes correlated features among observations of specified length or time range. However, in natural brain system, cognitive processes are affected by wider range of inputs taken from sensors. Therefore, in order to make machine capable of learning optimal range of input maps that the multiplicative unit takes features from, a reinforcement learning model is specified over time range to help the machine take different length of inputs sequences by solving the optimal value of the reward function defined by the model.

## IV. BI-TEMPORAL-RELATED MODEL

In Bi-related feature learning, machine extracts features from two related input maps. In order to combine the pair of two maps, a multiplicative unit is constructed to take the matrix multiplication of these maps as an input, and then is connected to a stack of hidden layers through k different kernels. Each hidden layer is constructed by sampling from a probabilistic distribution gained by convoluting the input with k kernels separately. The parameters of the model include k kernels and the biases associated with the hidden layers and the multiplicative unit respectively.

To describe the probabilistic distribution learned by model more explicitly, we redefine the traditional energy function of RBM **[5][6]** as

$$E(v_1,v_2,h) = -\sum_k \sum_{i,j}^{N_h} \sum_{r,s}^{N_w} W_{rs}^k (v_1 v_2)_{i+r-1,j+s-1} h_{ij}^k - \sum_k c_k h^k - b v_1 v_2 \quad (1)$$

where k is the index for hidden layers, n is the index for input maps, W represents the kernel matrix, and $N_h$ $N_w$ denotes the dimensions (rows and cols) of the hidden layers and of kernel matrix respectively. In **[5][6]**, the probabilistic model RBM based on energy above is defined by

$$P(v,h) = \frac{e^{-E(v,h)}}{Z},$$

$$P(v|h) = \frac{P(v,h)}{P(h)} = \frac{e^{-E(v,h)}/Z}{\sum_v e^{-E(v,h)}/Z} = \frac{e^{-E(v,h)}}{\sum_v e^{-E(v,h)}},$$

$$P(h|v) = \frac{P(v,h)}{P(v)} = \frac{e^{-E(v,h)}}{\sum_h e^{-E(v,h)}}$$

Then, we can derive the form of conditional distribution of hidden layers over the combined input maps as

$$P(h_{ij}^k = 1 | v_1, v_2) = sigmoid((W^k * (v_1 v_2))_{ij} + c_k)$$

which defines the distribution of the $ij^{th}$ element of the $k^{th}$ hidden layer conditioned on two input maps. To see this, we let $h_{xy}^z$ denote the $xy^{th}$ element of the $z^{th}$ hidden layer, and $h(-1)$ denote the other elements. If we define

$$A_{xy}^z(h) = -\sum_{r,s}^{N_w} W_{rs}^k (v_1 v_2)_{x+r-1, y+s-1} - c_z,$$

$$B(v, h(-1)) = -\sum_{k \neq z} \sum_{i,j \neq x,y}^{N_h} \sum_{r,s}^{N_w} W_{rs}^k (v_1 v_2)_{i+r-1, j+s-1} h_{ij}^k$$
$$- \sum_{k \neq z} c_k \sum_{i,j \neq x,y} h_{ij}^k - b v_1 v_2$$

, then it is easy to show that

$$P(h_{xy}^z = 1 | v) = \frac{e^{-A_{xy}^z(h)} e^{-B(v,h(-1))}}{e^{-B(v,h(-1))} e^{-A_{xy}^z(h) \cdot 1} + e^{-B(v,h(-1))} e^{-A_{xy}^z(h) \cdot 0}}$$
$$= sigmoid(-A_{xy}^z)$$
$$= sigmoid((W^z * (v_1 v_2))_{xy} + c_z)$$

where we used notation "*" to represent valid convolution, in which the last term can be viewed as a convolutional kernel. We can also write it in matrix form if we refine "sigmoid" as an operation on each matrix's element as

$$P(h^k = 1 | v_1, v_2) = sigmoid(W^k * (v_1 v_2) + c_k)$$

Similarly, the conditional distribution of the multiplicative unit is

$$P(v_1 v_2 = 1 | h) = sigmoid(\sum_k \overline{W}^k * h^k + b_2)$$

After concating it with the original one, we get the full generative multiplicative-unit. We denote the original and the generative multiplicative unit by $O, \overline{O}$. Then, by approximating the generative multiplicative unit as $\overline{O} \approx \overline{v_1} v_2 \approx v_1 \overline{v_2}$, we can use least squares to find the approximated version of each generative input maps.

$$\overline{v_1} = lstsq(v_2^T, \overline{O}^T)^T \quad (2)$$
$$\overline{v_2} = lstsq(v_1, \overline{O}) \quad (3)$$

where we have used "lstsq()" as the notation of least square method that solves the equation

$$Ax = B$$
$$x = lstsq(A, B)$$

the bar above O means the generative version of multiplicative unit, and "T" represents matrix transpose.

Ideally, we want to learn kernel and bias by maximizing the log likelihood in gradient ascent fashion given the conditional distributions above **[5][6]**. The gradient of the log likelihood for a CRBM based on energy model is

$$\frac{\partial Ln(P(\theta|v))}{\partial \theta} = \frac{\partial \ln(\sum_h e^{-E(v,h)})}{\partial \theta} - \frac{\partial \ln(\sum_{v,h} e^{-E(v,h)})}{\partial \theta}$$
$$= -\sum_h P(h|v) \frac{\partial E(v,h)}{\partial \theta} + \sum_{v,h} P(v,h) \frac{\partial E(v,h)}{\partial \theta} \quad \text{Unfo}$$
(4)

rtunately, computing this gradient involves an exponential number of terms. However, Hinton in **[6]** described a novel way, namely the so called contrastive divergence to approximate the gradient of (4), and then use this approximated gradient to update parameters via gradient ascent procedure. In contrastive divergence, for every iteration of updates, the multiplicative unit is sampled over hidden layers by m times. Take the $k^{th}$ kernel matrix for example, its gradient can be approximated as

$$\frac{\partial Ln(P(W|v))}{\partial W^k} = (v_1^{(0)} v_2^{(0)}) * P(h^k = 1 | v_1^{(0)}, v_2^{(0)})$$
$$- (v_1^{(m)} v_2^{(m)}) * P(h^k = 1 | v_1^{(m)}, v_2^{(m)}) \quad (5)$$

where the upper index over input layers means the original input maps and those sampled by m times respectively. The m-step CD procedure does converge as shown by Hinton **[6]**. Similarly, the gradients associated with bias can be written as

$$\frac{\partial Ln(P(\theta|v))}{\partial c_k} = P(h^k = 1 | v_1^{(0)}, v_2^{(0)}) - P(h^k = 1 | v_1^{(m)}, v_2^{(m)})$$
$$\frac{\partial Ln(P(\theta|v))}{\partial b} = v_1^{(0)} v_2^{(0)} - v_1^{(m)} v_2^{(m)} \quad (6)$$

To see that the gradients we get have the matrix form we want. We denote the dimension of multiplicative unit, kernel, and hidden layers by $N_V$, $N_W$, $N_H$. By the definition of convolution, $N_H = N_V - N_W + 1$. According to (5), the gradient of the kernel is gained by take the difference between two convolution, in which the hidden layer acts as a kernel. Therefore, by definition, the dimension of the gradient is $N_g = N_V - (N_V - N_W + 1) + 1 = N_W$. Similarly, it is easy to show that the dimension of bias's gradients are equal to that of bias. Thus, we verified that the gradients have the correct form.

Pseudocode of m-step constrastive divergence for training bi-related C-RBM is provided in algorithm 1.

Algorithm 1: Bi-temporal-related CRBM using m-step contrastive divergence

Initialize learning rate η = 0.2,

For t=0, 1, 2, 3 ...,T
    for i = 0, 1, 2, 3, ..., m
        Set multiplicative unit O(i) := $V_t(i) \times V_{t+1}(i)$
        For k = 0, 1, 2, 3, ..., K
            $H^k$ = Bernoulli(P($h^k$=1|O))
            If (i==0)
                Gradient(0)k = O(i) * P( $h^k$(t)=1|O(i) )
            If (i==m)

```
        Gradient(m)k=O(i) * P( h^k(t)=1|O(i) )
    End for
    V_t(i) = P( V_t(i)|h(t) )
    V_{t+1}(i) = P( V_{t+1}(i)|h(t) )
End for

For k = 0, 1, 2, 3, ..., k
    W^k = W^k + η(Gradient(0)k - Gradient(m)k)
    C_k = c_k + ( P(h^k=1|O(0) - P(h^k=1|O(m)) )
End for

b_t = b_t + V_t(0) - V_t(m)
B_{t+1} = b_{t+1} + V_{t+1}(0) - V_{t+1}(m)
    End for
End for
```

In the algorithm, index k represents time, and the second index i represents the step of CD. Figure.1 below shows the general structure of Bi-related model.

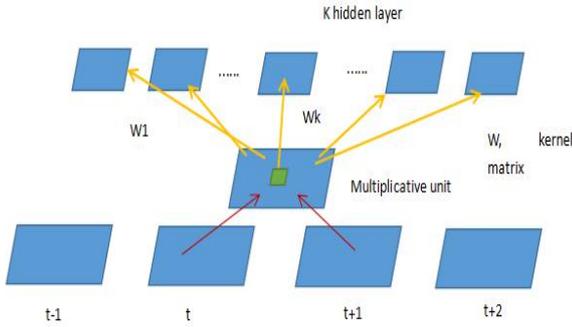

Figure.1: Bi-temporal-related model

## V. Tri-Temporal-Related Model

As we have discussed, bi-related model is sufficient to handle temporal learning task with the advantage brought by deep CRBM. However, in practice we need to extend bi-multiplicative unit to learn correlation from more than two maps. The reason for making that extension is that in many image recognition problems, more that two maps are correlated. Therefore, constructing hidden layers to extract relation between only two maps is inefficient and space-consuming. Also, in other cases where the rate at which the frames are sent into the unit might be too fast for the model to manipulate. For example, in cases where we may want to predict the objects' locations in a video based on features captured by temporal CRBM from inputs that are gained by cutting video input into pieces of frames, each of which represents a input map, the model does not have enough time to train network completely on two-related input maps before next pair of related maps shows up. Therefore, we have no options by to keep the rate of frames down. In other words, we need to slow down the rate at which video input is cut by frames into the model to give our machine more time to learn features from any pair of maps. However, this is extremely time-consuming.

In this section, we introduce a more general structure to address these problems. Recall that we have defined the term relational-order as the number of input maps that the hidden layers capture features from, we want machine to break the constrain of extracting features from limited number of maps by combining more related maps together, boosting up learning efficiency and reducing the unnecessary waste of learning space for parameters and hidden layers that are constructed during the training process. More specifically, in a typical tri-relational model, each multiplicative units takes account of three subsequent maps instead of two, and a stack of hidden layers being constructed to learn correlation among these three maps. Since the energy function and the conditional distribution in bi-related model do not hold for tr-related model, we need to slightly change the form of (1), (2), (3) to

$$E(v_1, v_2, v_3, h) = -\sum_k \sum_{i,j}^{N_h} \sum_{r,s}^{N_w} W_{rs}^k (v_1 v_2 v_3)_{i+r-1, j+s-1} h_{ij}^k - \sum_k c_k h^k - b v_1 v_2 v_3 \quad (7)$$

$$\overline{v_1} = lstsq((v_2 v_3)^T, \overline{O}^T)^T \quad (8)$$

$$\overline{v_2} = lstsq(v_1, lstsq(v_3^T, \overline{O}^T)^T) \quad (9)$$

$$\overline{v_3} = lstsq(v_1 v_2, \overline{O}) \quad (10)$$

Then we are able to train the model by Algorithm 1 with m-step constrastive divergence, in which for each step, input maps are sampled over conditional distributions (8) - (10). Figure.2 shows the structural difference between Bi-related and Tri-related model.

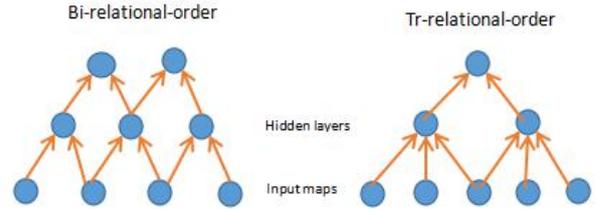

Figure.2: Bi-related model vs Tri-related model

## VI. Higher-Temporal-Related Model based on Reinforcement learning procedure

We then extend tr-temporal-related CRBM to higher-temporal-related CRBM with relational-order higher than three. Assume we are now using N-relational-order CRBM, we write conditional distributions of each input maps as

$$E(v_1, v_2, v_3, \cdots v_N, h) = -\sum_k \sum_{i,j}^{N_h} \sum_{r,s}^{N_w} W_{rs}^k (v_1 v_2 v_3 \cdots v_N)_{i+r-1, j+s-1} h_{ij}^k$$
$$- \sum_k c_k h^k - \sum_n b v_1 v_2 v_3 \cdots v_N \quad (11)$$

$$\overline{v_1} = lstsq((v_2 v_3 \cdots v_N)^T, \overline{O}^T)^T \quad (12)$$

$$\overline{v_2} = lstsq(v_1, lstsq((v_3 v_4 \cdots v_N)^T, \overline{O}^T)^T) \quad (13)$$

$$\overline{v_3} = lstsq(v_1 v_2, lstsq((v_4 v_5 \cdots v_N)^T, \overline{O}^T)^T) \quad (14)$$
$$\vdots$$
$$\overline{v_N} = lstsq(v_1 v_2 \cdots v_{N-1}, \overline{O}) \quad (15)$$

Then we use algorithm 1 to train the model by constructing a multiplicative unit O that learns features from N maps.

**Algorithm 2: Higher order temporal CRBM**

Initialize learning rate η = 0.2,
For t = 1, 1+N, 1+2N, ...
    for i = 0, 1, 2, 3, ..., m
        Set multiplicative unit O(i) := $V_t(i) \times V_{t+1}(i) ... \times V_{t+N}(i)$
        For k = 0, 1, 2, 3, ..., K
            $H^k(t)$= Bernoulli( P( $h^k(t)$=1|O ) )
            If (i==0)
                Gradient(0)k=O(i) P( $h^k(t)$=1|O(i) )
            If (i==m)
    Gradient(m)k=O(i)*P ($h^k(t)$=1|O(i))
        End for
        $V_t(i)$ = P( $V_t(i)$|h(t) )
        $V_{t+1}(i)$ = P( $V_{t+1}(i)$|h(t) )
        ...
        $V_N(i)$ = P( $V_N(i)$|h(t) )      Through equation (12) - (15)

End for

For k = 0, 1, 2, 3, ..., k
    $W^k$ = $W^k$ + η(Gradient(0)k - Gradient(m)k)
    $C_k$ = $c_k$ + (P($h^k(t)$=1|O(0) - P($h^k(t)$=1|O(m))
End for

$b_t$ = $b_t$ + $V_t(0)$ - $V_t(m)$
$b_{t+1}$ = $b_{t+1}$ + $V_{t+1}(0)$ - $V_{t+1}(m)$
...
$b_N$ = $b_N$ + $V_N(0)$ - $V_N(m)$

In the algorithm, index k represents time, and the second index i represents the step of CD.

As we have discussed in Section 3, one major problem faced by specific relational-order CRBM is the lack of generality. Once the relational-order is determined, the model is constrained to learn features from specific number of maps during training process. However, in many cases, sequences of input maps and their correlations are not the same. For example, in some input sequences, the number of strongly related maps might be three, but it may change to six later on. Therefore, it is reasonable to make relational order evolve through time. To do this, we first proved that the sub-sequence of a given sequence of input maps satisfies the sub-problem structure in dynamic programming. Then we develop a reinforcement learning procedure to learn the relational-order of each multiplicative unit by minimizing the reconstruction error of the whole sequence of inputs recursively through dynamic programming method that minimizes the reconstruction error of each sub-sequence input maps. Finally, we proved that the optimality is hold by this method.

We first denote a Markov decision process as (S, A, r), where "S" is the state-space, "A" is the action-space, and "r" represents the reward function [11][13]. r(s, a) is defined as the reward returned by taking action from state s. The task is to learn a policy $\pi$ that maps from the current state to an action. One obvious approach to determine the overall value of a policy is to evaluate the cumulative reward through that policy over time. More formally, cumulative reward following a policy $\pi$ from a given state s is defined as,

$$V^\pi(s_t) = \sum_{i=0}^{\infty} c^i r_{i+1}, \quad 0 \leq c < 1,$$ where c is a constant that determines the relative value versus immediate reward. More specifically, the importance of reward received at i time step is decreased exponentially by a factor $c^i$. Based on definitions above, the best policy we want the model to learn is the policy that gives the biggest cumulative reward. That is,

$$\pi^* = \arg\max_\pi V^\pi(s_t) \quad (16).$$

With that, the model can take the best action following the optimal policy defined above. Alternatively, we can base on cumulative reward function to choose "a" among actions as well. In that case, we redefine the best policy as

$$\pi^* = \arg\max_a \{r(s,a) + cV^*(s')\} = \arg\max_a \{Q(s,a)\} \quad (17)$$

where s' denotes the state after taking action a, and function Q is defined as the largest cumulative reward after taking action a at state s. By writing the relation between reward function and Q function more explicitly as

$$V^*(s) = \max_{a'} Q(s,a')$$

it is easy to see that Q can be defined as

$$Q(s,a) = r(s,a) + c \max_{a'} Q(s',a') \quad (18)$$

This recursive form of Q value function provides a basis for the algorithm that iteratively approximates value of Q efficiently [11]. [11][13] shows that the Q value gained by recursive method does converge to the optimal value.

To apply reinforcement-learning to our model, we view each relational-order as a state in "S". "A" contains three actions, "+1", "0", and "-1"(Here we assume that the relational-order can only change by one each time for simplicity). The goal is to minimize the average reconstruction error,

$$J(N) = \sum_n \sum_{i,j} \frac{1}{2} \left\| v_n^{(0)} - v_n^{(m)} \right\|^2 / (N * f(N)),$$

where the upper index (0) represents the original map, and (m) represent the one reconstructed by m times through constrastive divergence. The function of dominator N is to normalize reconstruction error so J of each relational-order has the same scale. Since we'd like to make model choose higher-order so that it can model a sequence of inputs with fewer number of multiplicative units and thereby reduce the number of parameters, we multiply N with an monotone increasing function f with respect to relational-order N. When N increase, f(N) will also increase, and therefore if reconstruction error of each input maps do not change by much, the J value would decrease. We than define the reward function as $R = -J$, so maximizing the reward is the same as minimizing the reconstruction error. In general procedure, we initialize all Q value to one, and choose the first relational order according to a gaussian distribution. After taking an action based on a probabilistic distribution defined as

$$P(a|s) = \frac{\exp Q(s,a)}{\sum_{a'} Q(s,a')},$$ update the current Q(s,a) by (18),

where a' denotes the action taken at the next state s'. Note that if the current state is two, then the model can only take the action "+1" or "0". Along this procedure, the model stores a table of Q value representing the reward at each relational-order state. Based on this table, the model can choose the relational-order itself that has the least reconstruction error at each training step by taking actions according to the distribution P(a|s). Also, we see that this brings the advantage of short-term memory, since the previously accumulative rewards added to Q will influence the state of the current training step. This also confirms the previous assumption that the relational-orders of nearby multiplicative units are also correlated. If the relational-order of current unit is 4, then that of the next unit would probably be around 4, instead of changing to 30 suddenly.

Reinforcement--learning procedure canhelp each multiplicative unit find its optimal relation-order along which minimizes the reconstruction error of the input maps. But, would minimizing the reconstruction error of each sub-sequences minimizes the reconstruction error of the whole sequence of input maps? We now show it indeed is. Assume that we have a sequence of input maps with length T. And we construct N multiplicative units, each of which has its relational-order $n_1, n_2, ,, n_N$. $J_T$ is defined as

$$J_T = J_{n_1} + J_{n_2} + \cdots + J_{n_N}$$

, which is the reconstruction error of the whole sequence. According to dynamic programming, we have

$$\min(J_T) = \min(J_{T-1} + \min J_{n_N}) \quad (19).$$

We now prove it is sufficient by giving

$$\begin{aligned} J_T &= J_{n_1} + J_{n_2} + \cdots + J_{n_N} \\ &= J_{T-1} + J_{n_N} \\ &\geq J_{T-1} + \min J_{n_N} \\ &\geq \min(J_{T-1} + \min J_{n_N}) \end{aligned}$$

Therefore, (19) does give the optimal reconstruction error. Then, using (19), we see that

$$\begin{aligned} \min(J_T) &= \min(J_{T-1} + \min J_{n_N}) \\ &= \min(J_{T-2} + J_{n(N-1)}) + \min J_{n_N} \\ &\vdots \\ &= \min J_{n_1} + \min J_{n_2} + \cdots + \min J_{n_N} \quad (20) \end{aligned}$$

This confirms the statement before that the optimal of J of the whole sequence can be gained by finding the optimal J of each sub-sequence. This is the core of dynamic programming, and in artificial intelligence it is called reinforcement learning, which is what we have introduced earlier. Thus, we use our reinforcement-learning procedure to train the high-temporal-related model at each learning epoch.

Below is the pseudocode

Algorithm 3: Higher-related model

Initialize all Q(s,a) and r(s,a) to one

Initialize all weights and bias by gaussian distribution with mean 0 and variance 1.

Set initial relational order s to arbitrary number that is greater than 1.

For epoch e = 0, 1, ,,, E
For t = 0, 1, 2, ...,T
  Run algorithm 2 with relational-order n=S(t)
  Choose "a" according to P(a|S(t))

  If "a" is "-1" & S > 2
    Update state, S(t+1) := S(t)-1
  If "a" is "0"
    Update state, S(t+1) := S(t)
  If "a" is "+1"
    Update state, S(t+1) := S(t)+1

Run algorithm 2 with relational-order N= S(e)

  Set r(S(t), a) = -$J_N$ (set the reward at state S taking action a to the minus of reconstruction error)

  Update the value of Q(S(t), a) by equation (18)
End for
End for

### Ⅶ. Experiments

This section focus on experiments that demonstrate the performance of temporal-related CRBM in practice. We will use the famous MINIST-dataset(http://yann.lecun.com/exdb/mnist/). At each training process, we give the model a sequence of digit input maps. And than we apply TD-CRBM to learn the matrix transformations among the maps. Finally, we use the matrix transformations we learned to produce generative maps of each digit inputs and then measure the reconstruction error to show model performance.

#### A. Performance of Bi-Temporal-Related model in MINIST-dataset

We first test model's ability to capture simple transformation such as shifts and rotation of a single digit. We construct a bi-temporal model, which consists of a multiplicative unit with relational-order 2, each of which connects to a stack of K hidden layers. We then give the model 10 sequences of input maps of length 100, each of which consists of 100 digit maps of different angles towards the axis. We train the model following algorithm 1 and store each stack of hidden layers as feature maps that describe the correlation between each pair of digit inputs. After each training epoch is finished, we calculate the reconstruction error between the generative digit inputs and the original ones. We see that as the training epoch increases, the reconstruction error drops out exponentially, which confirms the efficiency and ability of bi-related model. Figure.3 shows the performance of Bi-related model in the task of learning transformation between pair of MINIST-digit maps.

#### B. Higher-related with Q-learning VS Bi- & Tri-related

As our discussion in previous sections, higher-related model is more practical and can reduces the amount of parameters that must be learned. In this section, we implemented an experiment that shows different performance of higher, tri, and bi-temporal-related models.

Again, we give each model 10 sequences of maps, each of which consists of 100 digit maps of the same number digit. We constraint on the amplitude of the transformation between pair of inputs. More specifically, the rotation from the last map do not exceed 20 degrees, and the shift of the centroid do not exceed 2. We then use algorithm 1 to train bi-related model, algorithm 2 with relational-order 3 to train tri-related model, and algorithm 3 with initial order 5 to train higher-related model.

Figure.3 and .4 summarizes the results of the experiment, from which we can evaluate the performance of three different models.

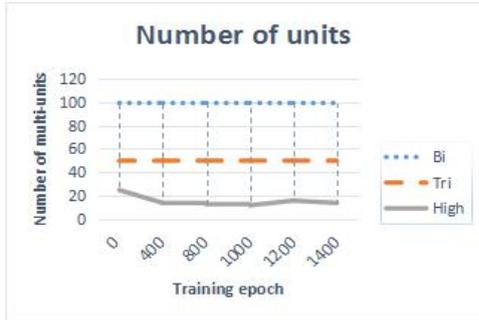

| Model \ Epoch | 0 | 400 | 800 | 1000 | 1200 | 1400 |
|---|---|---|---|---|---|---|
| Bi | 99 | 99 | 99 | 99 | 99 | 99 |
| Tri | 50 | 50 | 50 | 50 | 50 | 50 |
| Higher | 25 | 14 | 13 | 12 | 16 | 14 |

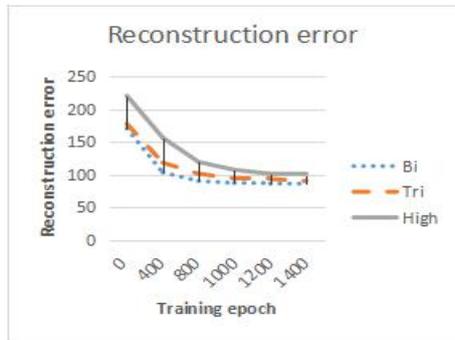

Figure.3: Performance of the three models when nearby inputs are highly correlated

We now throw away the constraint on transformation to see how higher-related model behaves.

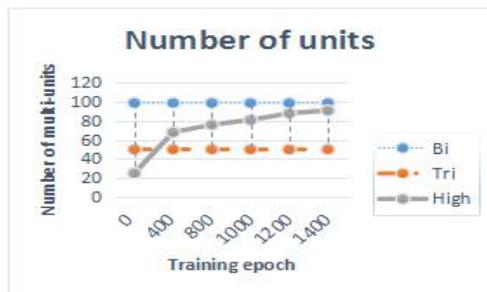

| Model \ Epoch | 0 | 400 | 800 | 1000 | 1200 | 1400 |
|---|---|---|---|---|---|---|
| Bi | 99 | 99 | 99 | 99 | 99 | 99 |
| Tri | 50 | 50 | 50 | 50 | 50 | 50 |
| Higher | 25 | 68 | 76 | 81 | 88 | 91 |

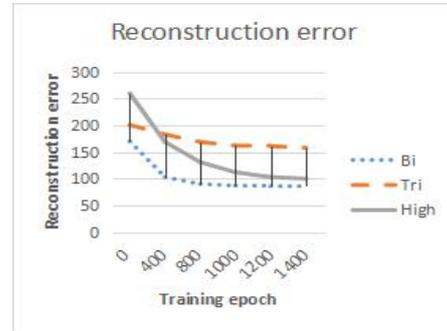

Figure.4: Performance of the three model when there is no constraint on transformation

### C. Results and conclusion

We first analyze bi-related model. We see from Figure.3 and Figure.4 that performance of bi-related model are the same in each case. This is because 2rd relational-order multiplicative unit is the most basic one. And the feature map learned by each unit can well describe the relationship between pair of input maps, no matter how much one map transform into another.

In tri-related model, however, we see even though it has roughly the same training epoch to that of bi-related, it does not perform well when there is no constraint on the transformation between each pair of maps. And the reconstruction error is still above "170" even after 1400 training epochs. This is true because when there is no limitation on transformation, the least number of highly related maps is only two. Each multiplicative unit can only learn features from two maps with one transformation. When a $3^{rd}$ relational-order unit is constructed to learn transformation from three maps with two far different transformation, the error goes up. For example, consider a sequence of three maps, the first map rotates 90 degrees clockwise into the second map, and the second rotates 120 degrees counterclockwise into map three. In this case, bi-related model constructs two multi-units, each of which learns the feature map of one rotation. The Tri-related model, however, will not perform well because one multi-unit can not learn two different transformations at the same learning epoch.

We see that higher-order model with reinforcement--learning procedure is more flexible. It can choose the optimal relational-order for each multi-units to give the least reconstruction error. In Figure.3, when each map is highly related, higher-order model constructs multi-units with relational-order higher than 5, thereby reducing the number of parameters dramatically. Then in Figure 4, when there is no constraint on transformation, we see that even though its error during early steps of training is far higher that those of other two models, high-order model can adjust the relational-order to the optimal that gives the least reconstruction error. After 400 training epochs, it outperforms tri-related model. After 1400 training epochs, it reduce its reconstruction error to roughly the same scale to that of bi-related model.

## VIII. Conclusion: Limitation & Further research

In this article, we introduced a CRBM structure that models temporal-dependent input sequences by constructing multiplicative units and hidden layers to learn features of correlation among related input maps. We then combined this model with reinforcement--learning procedure. By doing so, we made our model capable of learning the optimal relational-order for each multiplicative units to give the least amount of errors. From the experiment, we see that the model with reinforcement--learning procedure is more flexible on choosing the relational-order. And it reconstructs the generative maps with the same accuracy to that of bi-related model in the long-term.

However, reinforcement--learning strategy increases the complexity of computation and the amount of training time dramatically. We see that it takes about 1400 training epochs for the model to reduce its reconstruction error down to 100. Its performance only catches up with that of bi-related model in the long-term. Therefore, further researches to reduce the training period are needed.

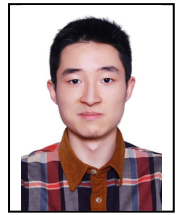
**Zizhuang Wang** is a Chinese high school student entering his 12th grade in year of 2016 . After finishing all school courses with a perfect GPA and won multiple school scientific competitions at 11th grade, he started to focus on researches of artificial intelligence and mathematics. Wang's major research interest lies in machine learning, neural networks, bayesian inference, applied mathematics, and quantum computation. (email: 1012125144@qq.com ;)
Web: http://kingofspace0wzz.github.io/